\documentclass[12pt]{elsarticle}
\usepackage[whole]{bxcjkjatype}

\usepackage{amssymb}
\usepackage{amsmath}
\usepackage{url}
\usepackage{graphicx}
\usepackage[ruled,vlined]{algorithm2e}
\usepackage{hyperref}
\hypersetup{
    colorlinks=true,
    linkcolor=blue,
    filecolor=magenta,
    urlcolor=cyan,
}
\journal{Frontiers in Artificial Intelligence}

\begin{document}

\begin{frontmatter}

\title{Who Speaks Next? Multi-party AI Discussion Leveraging the Systematics of Turn-taking in Murder Mystery Games}

\author{Ryota Nonomura}
\ead{nono@speech-lab.org}
\author{Hiroki Mori\corref{cor1}}
\ead{hiroki@speech-lab.org}
\affiliation{organization={School of Engineering, Utsunomiya University},
            addressline={7-1-2, Yoto}, 
            city={Utsunomiya},
            postcode={321-8585}, 
            state={Tochigi},
            country={Japan}}
\cortext[cor1]{Corresponding author}
\begin{abstract}
Multi-agent systems utilizing large language models (LLMs) have shown great promise in achieving natural dialogue. However, smooth dialogue control and autonomous decision making among agents still remain challenges. In this study, we focus on conversational norms such as adjacency pairs and turn-taking found in conversation analysis and propose a new framework called ``Murder Mystery Agents'' that applies these norms to AI agents' dialogue control. As an evaluation target, we employed the ``Murder Mystery'' game, a reasoning-type table-top role-playing game that requires complex social reasoning and information manipulation. In this game, players need to unravel the truth of the case based on fragmentary information through cooperation and bargaining. The proposed framework integrates next speaker selection based on adjacency pairs and a self-selection mechanism that takes agents' internal states into account to achieve more natural and strategic dialogue. To verify the effectiveness of this new approach, we analyzed utterances that led to dialogue breakdowns and conducted automatic evaluation using LLMs, as well as human evaluation using evaluation criteria developed for the Murder Mystery game. Experimental results showed that the implementation of the next speaker selection mechanism significantly reduced dialogue breakdowns and improved the ability of agents to share information and perform logical reasoning. The results of this study demonstrate that the systematics of turn-taking in human conversation are also effective in controlling dialogue among AI agents, and provide design guidelines for more advanced multi-agent dialogue systems.
\end{abstract}



\end{frontmatter}

\section{Introduction}
\label{sec:Introduction}

The emergence of large language models (LLMs) has dramatically enhanced the capabilities of AI agents. With the advent of LLMs such as GPT-3, GPT-4, and LLaMA, we have witnessed the achievement of human-comparable or superior performance across various tasks, including text generation, question-answering, and summarization \cite{Brown2020,OpenAI2023,LLAMA,LLAMA2}. The development of AI agents based on these LLMs has gained significant momentum, with promising applications spanning diverse domains such as customer service \cite{Rome2024}, educational support \cite{Jeon2023,Hu2024Education,Zhang2024Education}, and creative work assistance \cite{OpenAI2022,Claude35}.

Of particular interest is whether AI agents can exhibit social behaviors similar to those of humans. Previous studies have employed various approaches to observe the social behaviors of LLM-based agents. For instance, Park et al. \cite{Park2023} conducted virtual daily life simulations, analyzing the behavioral patterns of 25 AI agents and their impact on a simulated society. Their study observed information sharing between agents and the formation of novel relationships.

Meanwhile, Lan et al. \cite{Lan2023} conducted research evaluating social interaction capabilities through multi-agent conversations in the board game Avalon, which requires cooperation and deception among multiple agents. Their study proposed a framework that enables AI agents to make strategic decisions based on previous gameplay experiences, reporting observations of social behaviors such as leadership and persuasion.

In social interaction, verbal communication plays a central role. Previous studies on the application of LLMs have also revealed that enabling AI agents to chat with each other is an effective approach. Qian et al. \cite{Qian2023} demonstrated that a chat chain between an instructor and an assistant is effective for completing various subtasks in the workflow of software development. Gu et al. \cite{Gu2024} proposed a simulation framework for group chats among AI agents, reporting that multifaceted emergent behavior was observed during role-playing scenarios. Wu et al. \cite{Wu2023} proposed a platform for LLM applications that supports interaction between LLMs, humans, and tools, where group chats among AI agents are facilitated.

However, text chats are significantly different from human-to-human conversations. It has been claimed that text chat is incoherent, especially due to the lack of interaction management such as simultaneous feedback, which leads to disruption and breakdown of turn-taking and topic management \cite{Herring1999}. Most AI chat systems employ an even simpler turn-taking model: sending text input from the user initiates the turn transition. This framework does not reflect the properties that human conversation has. For example, chat AIs cannot actively offer topics, initiate conversations, remain silent when other participants are to speak, or withhold from speaking. 

Turn-taking plays a crucial role especially in multi-party conversations, yet there have been relatively few studies on such conversation by AI agents. In order to handle multi-party conversations, the problem of selecting the next speaker arises. In the AutoGen platform \cite{Wu2023}, an automatic next-speaker selection mechanism is implemented, where an LLM agent estimates the next speaker's role based on the history of the speaker's role and utterances. However, Bailis et al. \cite{Bailis2024} pointed out that while this approach is potentially effective, it lacks autonomy for individual agents. Instead, they proposed a dynamic turn-taking system where agents express their desire to speak by bidding.

As Bailis et al. \cite{Bailis2024} argued, allowing agents to autonomously determine the speaking order could be key to AI agents playing their own social role and having a fruitful conversation. At the same time, however, the order of speaking should not be determined solely by the agents' will. Sociologists who pioneered conversation analysis devised a concept of adjacency pairs \cite{Sacks1973} as the basic unit of utterance sequences. An adjacency pair is a two-part exchange in which the second utterance is functionally dependent on the first. Such functional binding is called conditional relevance \cite{Schegloff1968}. When the current speaker addresses a question to another one, the addressee is not only obligated to take the turn, but also to speak something relevant to the question. In multi-party conversations, the first pair part of adjacency pairs often involves this ``current speaker selects next'' technique \cite{Sacks1974}.

Therefore, the research question here is whether introducing turn-taking systematics such as adjacency pairs, discovered in the research field of conversation analysis, into the next-speaker selection mechanism will have the effect of making LLM-based multi-agent conversations more natural and efficient. Schegloff \cite{Schegloff1990} argued that organization of sequences in turn-taking systematics such as adjacency pairs is the source of coherence in conversation. If so, introducing such a conversational norm into conversations by AI agents is expected to improve the coherence of conversation.

To address this research question, we developed Murder Mystery Agents (MMAgents), a system where multiple AI agents play a deductive tabletop role-playing game called Murder Mystery. MMAgents consists of a self-selection mechanism for autonomous utterances and a next-speaker selection mechanism that detects the first part of adjacency pairs using LLMs to determine the next speaker.

\section{Background}
\label{sec:Background}
\subsection{LLM-Based Agents}
\label{sec:LLM-Based Agent}
With the advancement of large language models (LLMs), numerous LLM-based agents have been proposed \cite{Wang2023, Wu2023, Significant_Gravitas_AutoGPT, BabyAGI}. These autonomous agents, built upon foundational models such as GPT-3 and GPT-4, are capable of executing complex tasks, engaging in assistant-like dialogue, and making decisions.
 
The applications of LLM-based agents span diverse domains, including software development \cite{Qian2023}, gaming environments \cite{Hu2024}, and economic simulations \cite{Horton2023}. Of particular interest are multi-agent systems involving multiple agents. The CAMEL framework \cite{Li2023} demonstrates how agents with distinct roles can collaborate to solve problems. Additionally, research on the Avalon Game \cite{Lan2023} simulates complex social interactions, including cooperation and conflict between agents.

Furthermore, research on AI agents' social behavior, particularly interaction through conversation, continues to evolve. These studies investigate the ability of multiple agents to participate in group chats and discussion scenarios, generating conversations that closely resemble human-to-human interactions \cite{Junprung2023, Gu2024}. Agents have been shown to possess capabilities such as memory, reflection, and planning, enabling more human-like dialogue \cite{Park2023}. This approach contributes to understanding the mechanisms of information exchange and cooperative behavior among agents, potentially offering insights into emergent behaviors in human society.

Research aimed at enhancing LLM-based agents' capabilities is also being actively pursued. For instance, SelfGoal \cite{Yang2024} proposes automatic generation and updating of sub-goals to achieve high-level objectives. Chain-of-thought prompting \cite{Wei2022} significantly improves performance on complex reasoning tasks by generating intermediate thought processes. Moreover, ReAct \cite{Yao2022} proposes an approach alternating between reasoning and action, enhancing agents' ability to solve problems incrementally while interacting with their environment.

However, there have been relatively few studies focusing on multi-party conversations among LLM-based agents. Most existing research deals with one-to-one interactions or simplified turn-taking mechanisms, failing to address the natural flow of conversation that occurs in groups of three or more participants.

\subsection{Turn-taking}
\label{sec:Turn-taking}
In human conversation, there exists a fundamental constraint where typically only one person speaks at a time. This constraint stems from the physical limitations of speech communication, as simultaneous speech by multiple participants leads to interference, making comprehension difficult.
For efficient communication, speakers must smoothly alternate turns while minimizing silent intervals between utterances. To meet this requirement, humans have naturally developed turn-taking systems through social interaction.

Turn-taking, where dialogue participants take turns to speak, forms the foundation of smooth communication. Through analysis of spontaneous conversation recordings, conversation analysts like Sacks, Schegloff, and Jefferson systematically described this phenomenon and identified the following rules \cite{Sacks1974}:
\begin{enumerate}
\item If the current speaker designates the next speaker by using a `current speaker selects next' technique (e.g., at the first pair part of an adjacency pair \cite{Sacks1973}), the selected participant has both the right and obligation to become the next speaker. (Current Speaker Selects Next)
\item If the current speaker does not designate the next speaker, other participants can spontaneously initiate speech. (Self-Selection)
\item If no one begins speaking, the current speaker can continue.
\end{enumerate}
Unlike dyadic conversations where speaker and listener roles are clearly defined, multi-party conversations involve multiple participants, necessitating the use of gaze direction and verbal addressing to designate the next speaker \cite{Sacks1974}.
Adjacency pairs, the basic units of conversation, consist of paired utterances such as [question-answer] and [invitation-acceptance/rejection]. The initial utterance is referred to as the first pair part, and the responding utterance as the second pair part. First pair parts like ``I'd like to purchase this item (request)'' generate an obligation for a specific type of second pair part (in this case, ``Certainly (acceptance)'' or ``We're sold out (rejection)''). An inappropriate second pair part or lack of response suggests either a communication error or implies a reason for the inability to respond.

Humans dynamically create conversations as collaborative acts among participants using this turn-taking system. In contrast, current AI agents struggle to autonomously engage in such flexible and immediate interactions. Therefore, implementing turn-taking mechanisms in AI agents may enable more natural and smooth dialogue.

\subsection{Murder Mystery}
\label{sec:Murder Mystery}
Murder Mystery is a reasoning-type table-top role-playing game in which players play the roles of characters within a story, aiming to either identify the murderer or, if playing as the murderer, to avoid detection. The game's progression heavily relies on players sharing information through conversation, including evidence gathered from crime scene investigations and character-specific knowledge.
Furthermore, Murder Mystery assigns different missions to each player. Players may need to cooperate or deceive others to accomplish these missions. This requires not merely intelligence but also human-like social behaviors such as teamwork, persuasion, negotiation, and deception. Successfully replicating these behaviors in AI agents could lead to significant advances in artificial intelligence research.

There has been one attempt to make AI agents play Murder Mystery games \cite{Junprung2023}. In this prior research, a detective agent poses the same questions to five agents, including the murderer. After all five responses are collected, the detective agent responds and asks another question. This process is repeated \(N\) times, after which the detective agent attempts to identify the murderer. This approach is termed ``one-to-many simulation.'' While the simulation successfully identifies the murderer, this method does not accurately reflect real Murder Mystery gameplay, where all players except the murderer must develop their own theories to identify the murderer. While this approach is referred to as ``many-to-many,'' it could not be implemented due to OpenAI's input token limitations. Therefore, this research aims to develop an agent framework capable of either reasoning or concealing information about the murder through autonomous conversation, similar to human players.

\section{Conversational Agents Simulating Human Multi-party Conversation}
\label{sec:Conversational Agents Simulating Human multi-Party Conversation}
Building upon the characteristics of Murder Mystery games discussed in Section \ref{sec:Murder Mystery}, this section details the design philosophy and technical components of MMAgents (Murder Mystery Agents), a system developed to facilitate autonomous game progression. MMAgents is designed to simulate multi-party human conversations, enabling multiple AI agents to not only cooperate but also engage in complex conversations involving competition and bargaining to advance the Murder Mystery game.
\subsection{Component}
\label{sec:Component}
\subsubsection{Character Setting}
\label{sec:CharacterSetting}
In Murder Mystery games, before the game begins, the game master provides players with character sheets. Each character sheet contains information necessary for players to portray their characters, including background, personality, objectives, and actions on the day of the incident. Players read and understand this information and play the character to talk and explore.

\begin{figure}[tbp]
  \centering
  \includegraphics[scale=0.5]{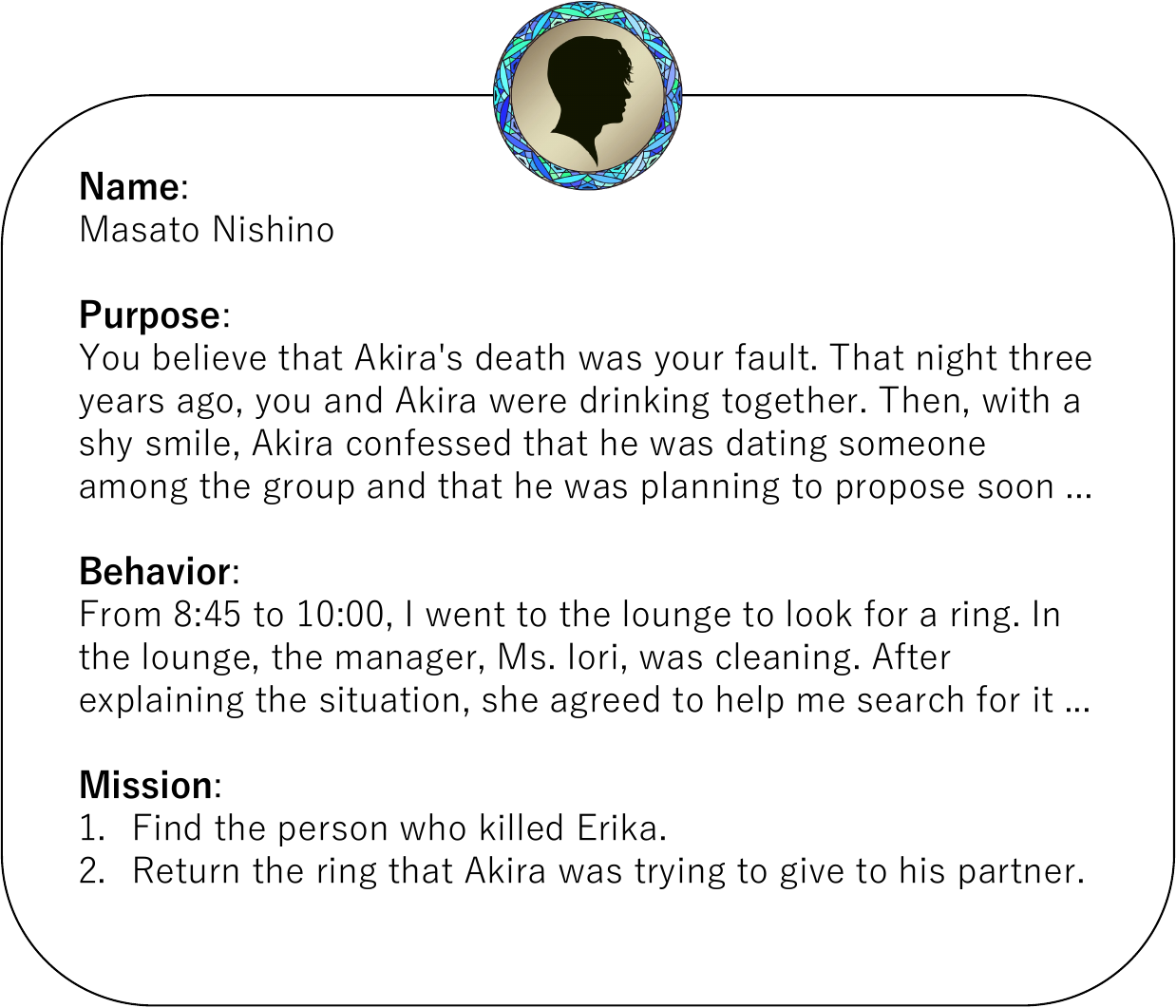}
  \caption{Example of character information. The original text is in Japanese. The same applies hereafter.}
  \label{fig:character information}
\end{figure}

The approach of having LLMs roleplay characters and evaluating their performance has been reported in several studies \cite{Shanahan2023,Shao2023,Wang2023RoleLLM,Lu2024}. As shown in Figure \ref{fig:character information}, MMAgents structures each agent's prompt beginning with the character's name, followed by descriptions of their objectives, actions, and missions to accomplish. For example, the character Masato Nishino's information includes crucial background details such as memories of his close friend Akira who passed away three years ago, and romantic feelings expressed that night. The information also includes specific incident-related actions, such as his behavior in the lounge the previous day and conversations with the inn's manager. Furthermore, character-specific missions are established, such as ``finding Erika's murderer'' and ``returning the ring that Akira intended to give to his lover.''

In this way, each agent is provided with character information containing distinct backgrounds and objectives, which guides their decision-making and dialogue. Only surface-level information about other characters is shared, and this information asymmetry implements the elements of information gathering and strategic interaction inherent in Murder Mystery games.

\subsubsection{Memory}
\label{sec:memory}
For LLM-based agents, memory management mechanisms are crucial components for generating more natural and consistent responses in user interactions \cite{Zhang2024,Zhong2023,Modarressi2023}. This is equally important in agent-to-agent dialogue \cite{Park2023}. To create systems like Murder Mystery, where multiple agents engage in complex discussions over extended periods, it is essential to appropriately store past statements and acquired information, and recall them at necessary moments.

Drawing inspiration from human memory systems, this research manages agents' memory across three distinct layers. First, there is a memory named \emph{History} that is shared by all agents, which maintains the past $k$ turns of conversation as shown in Equation (\ref{formula:history}). History is used to maintain conversational context and track recent dialogue flow.
\begin{equation}\label{formula:history}
  \textit{history} = \{u_{n-k+1}, u_{n-k+2}, ..., u_n\},
\end{equation}
where $u_i$ represents the $i$-th utterance.

Second, each agent maintains a short-term memory, named \emph{shortTermHistory}. This consists of a history of thoughts generated by the think() function detailed in Section \ref{sec:think()}, and maintains agent-specific policies and intentions, as shown in Equation (\ref{formula:shortTermHistory}),
\begin{equation}\label{formula:shortTermHistory}
  \textit{shortTermHistory} = \{t_{n-k+1}, t_{n-k+2}, ..., t_n\},
\end{equation}
where $t_i$ represents the $i$-th thought. The shortTermHistory enables agents to maintain consistency in their reasoning and intentions.

\begin{figure}[tbp]
  \centering
  \includegraphics[scale=0.7]{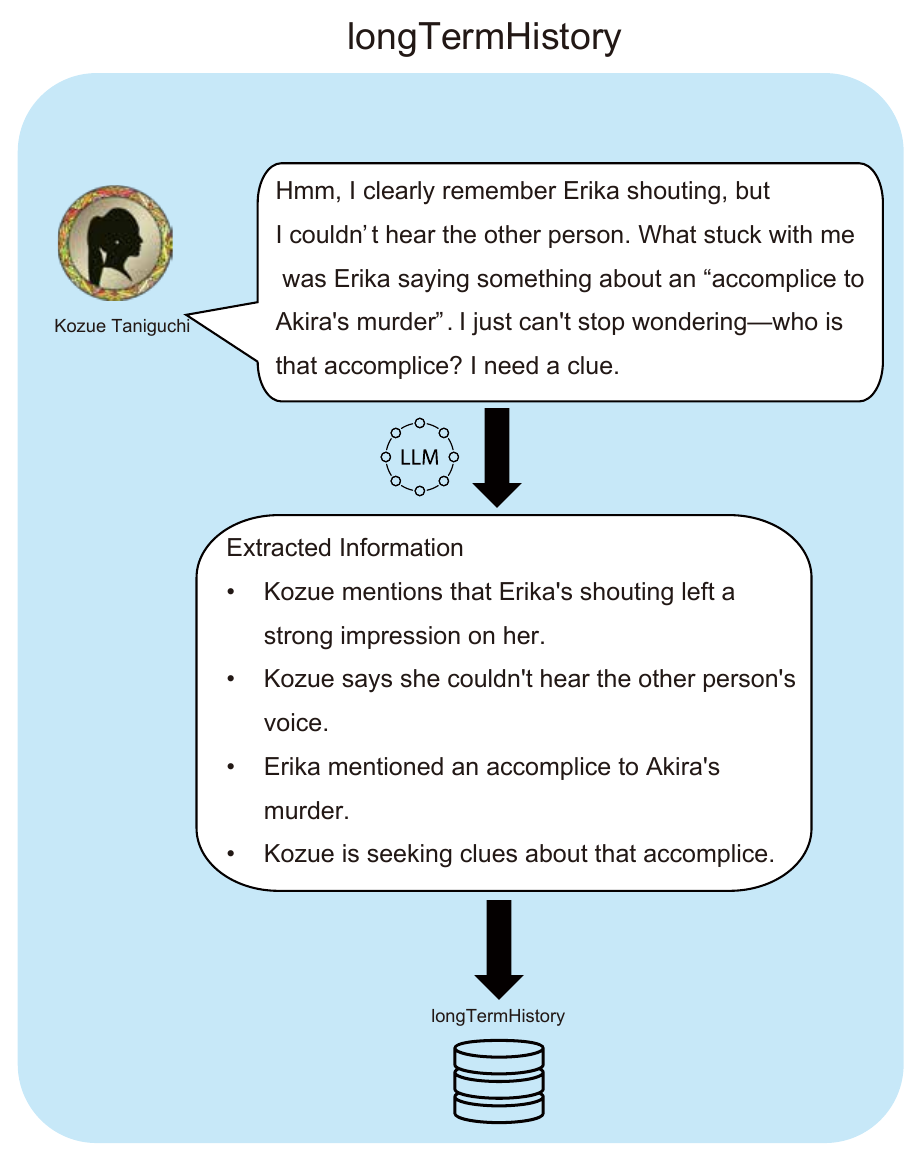}
  \caption{Example of normalizing utterances into knowledge or information and storing in longTermHistory.}
  \label{fig:longTermHistory}
\end{figure}

Furthermore, each agent maintains a long-term memory, named \emph{longTermHistory}, in which utterance content is normalized using LLMs, and important knowledge and information is extracted and stored in a database, as formulated in Equation (\ref{formula:longTermHistory}). Figure \ref{fig:longTermHistory} demonstrates the process of information extraction and normalization in longTermHistory. This example illustrates the process of extracting important information from unstructured speech text by Kozue Taniguchi and storing it as structured knowledge. This normalization process facilitates later retrieval and reference by extracting important facts and information from unstructured text in a bullet-point format.
\begin{equation}\label{formula:longTermHistory}
  \textit{longTermMemory} = \{{k}_{1}, {k}_{2}, \ldots\}
\end{equation}
When generating new utterances, the previous utterance $u_{t-1}$ is converted into an embedding vector $E(u_{t-1})$, and the cosine similarity shown in Equation (\ref{formula:cos sim}) is calculated with each vector $E(k_{i})$ of the embedded knowledge stored in longTermHistory to retrieve relevant past memories.
\begin{equation}\label{formula:cos sim}
  \cos(E(u_{t-1}), E(k_{i})) = \frac{E(u_{t-1}) \cdot E(k_{i})}{|E(u_{t-1})| |E(k_{i})|}
\end{equation}
The calculated similarities are sorted in descending order, and normalized knowledge ($k_i$) corresponding to the top $l$ vectors is selected. This enables efficient recall of past memories relevant to the current context, which agents can utilize for reasoning and utterance generation.
These three layers of memory systems each have different time scales and purposes. History maintains the flow of recent conversations, shortTermHistory retains each agent's thought processes, and longTermHistory stores important facts and information. By incorporating these memories into prompts, agents can generate contextually appropriate utterances and maintain consistent conversations.
\subsection{Turn-taking System}
\label{sec:Turn-taking System}
The turn-taking system is potentially a crucial element for achieving natural dialogue among multiple agents. In conventional multi-agent dialogue systems, speaking turn was often predetermined or randomly assigned. In this research, based on Sacks et al.'s conversation analysis theory discussed in Section \ref{sec:Turn-taking}, we implemented two characteristic turn-taking mechanisms from natural human conversation in MMAgents: ``Self-Selection'' and ``Current Speaker Selects Next''. This enables natural turn-taking that reflects the agents' personalities and intentions. The pseudocode for this algorithm is shown in Algorithm \ref{Turn-taking Algorithm}. This subsection details the important modules of the turn-taking algorithm.
\begin{algorithm}\label{Turn-taking Algorithm}
\DontPrintSemicolon
\caption{Turn-taking system}
 \textit{nextSpeaker} $\leftarrow$ nil \;
 \While{\textit{true}}{
     \textit{currentSpeaker} $\leftarrow$ \textit{nextSpeaker} \;
    \For{\textit{agent} \KwSty{in} \textit{agents}}{
        \textit{thought}, \textit{action}, \textit{importance} $\leftarrow$ \textit{agent}.think() \;
        \textit{agent}.thought $\leftarrow$ \textit{thought} \;
        \textit{agent}.action $\leftarrow$ \textit{action} \;
        \textit{agent}.importance $\leftarrow$ \textit{importance} \;
    }
    \If{\textit{currentSpeaker} \KwSty{is} nil}{
        \textit{currentSpeaker} $\leftarrow$ selectMostImportant(\textit{agents}) \;
    }
    \For{\textit{agent} \KwSty{in} \textit{agents}}{
        \eIf{\textit{agent} \KwSty{is} \textit{currentSpeaker}}{
            \textit{utterance} $\leftarrow$ \textit{agent}.speak() \;
            \textit{agent}.shortTermHistory.append(\textit{utterance}) \;
        }{
            \textit{agent}.shortTermHistory.append(\textit{agent}.thought) \;
        }
    }
    history.append(\textit{utterance}) \;
    longTermHistory.append(knowledgeNormalization(\textit{utterance})) \;
    \textit{nextSpeaker} $\leftarrow$ detectDesignation(\textit{utterance}) \;
}
\end{algorithm}

\subsubsection{think()}
\label{sec:think()}
At the beginning of each turn, agents execute an action called think(). Based on the provided character data, think() generates thought, which represents the plan for the next utterance or action aimed at achieving their mission. Simultaneously, it decides whether to take the action of ``speak'' or ``listen''. This selection is implemented with the assumption that it is determined by considering other agents' utterances and the urgency of their own thought content. Furthermore, it outputs an importance as an integer from 0 to 9. This value is designed to reproduce the Self-Selection mechanism in conversation and is presumed to be determined based on factors such as relevance to the mission, consistency with current conversational context, urgency of the utterance content, and character personality.

\begin{figure}[tbp]
  \centering
  \includegraphics[scale=0.5]{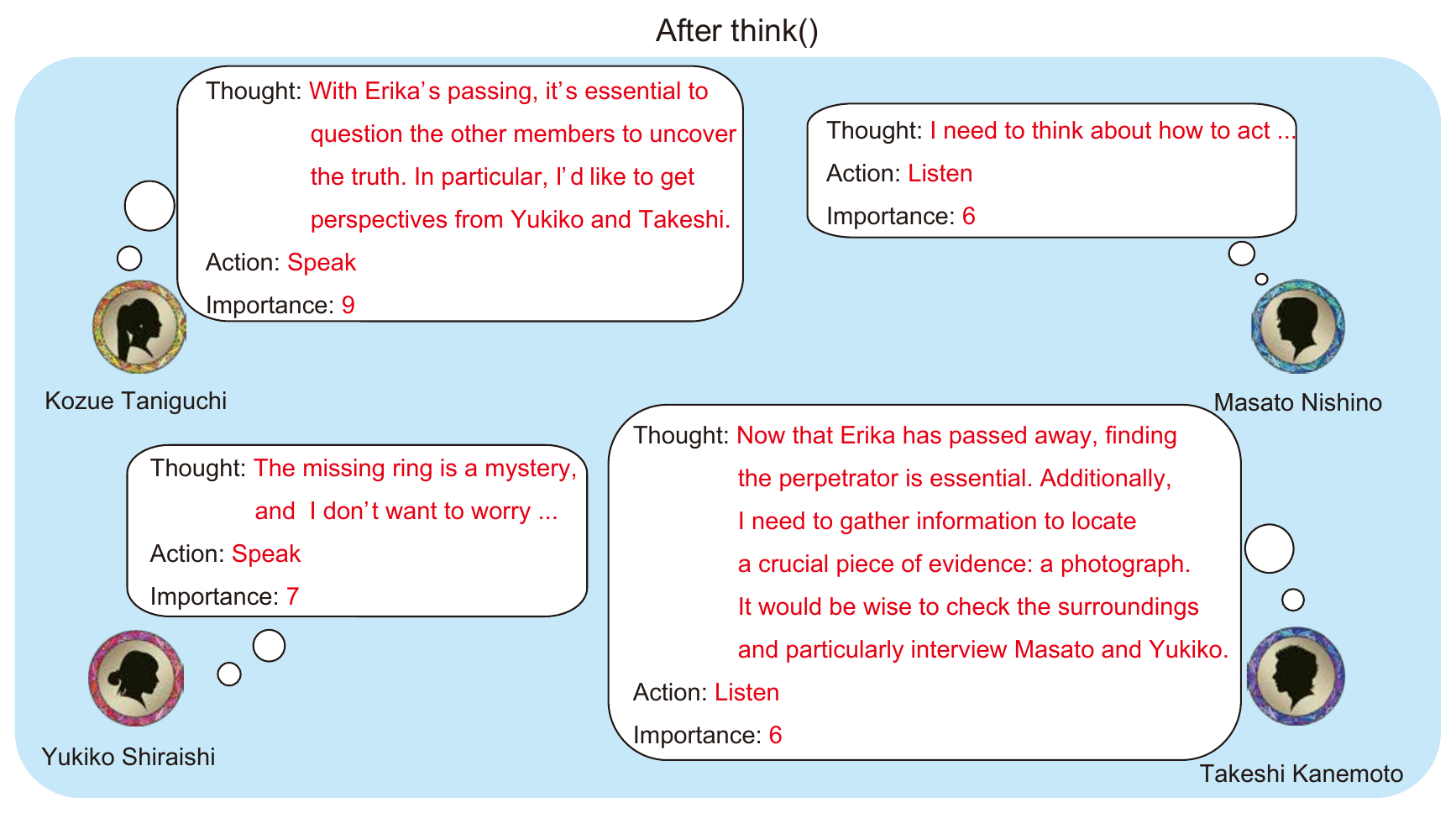}
  \caption{Example output of think().}
  \label{fig:think()}
\end{figure}

Figure \ref{fig:think()} shows an example where four agents execute think(). In this example, Kozue Taniguchi and Yukiko Shiraishi chose ``speak'', with Kozue Taniguchi in particular outputting a high importance value. This suggests that Kozue Taniguchi judged her utterance to be significant for the conversation's development.
\subsubsection{selectMostImportant()}
\label{sec:selectMostImportant()}
The selectMostImportant(agents) is a speaker selection algorithm that implements the Self-Selection mechanism. This algorithm processes differently based on the number of agents who have selected ``speak.'' When only one agent selects ``speak'', that agent naturally becomes the speaker. This is the simplest case of Self-Selection. Conversely, when multiple agents select ``speak'', their importance values are compared, and the agent with the highest value becomes the speaker. This represents the turn-taking systematics of ``the first person to start speaking becomes the speaker'', expressed numerically through importance values. In cases of tied importance values, random selection is used to represent the uncertainty of turn-taking in actual conversations. Furthermore, when all agents select ``listen,'' the previous speaker continues speaking. This implements the turn-taking systematics that ``when the current speaker does not select the next speaker, they retain the right to continue speaking''. However, in the first turn at the start of the dialogue, the speaker is determined randomly. In the example shown in Figure \ref{fig:think()}, although both Kozue Taniguchi and Yukiko Shiraishi selected ``speak'', Kozue Taniguchi is chosen as the next speaker due to her higher importance value.

\subsubsection{speak()}
\label{sec:speak()}
\begin{figure}[tbp]
  \centering
  \includegraphics[scale=0.65]{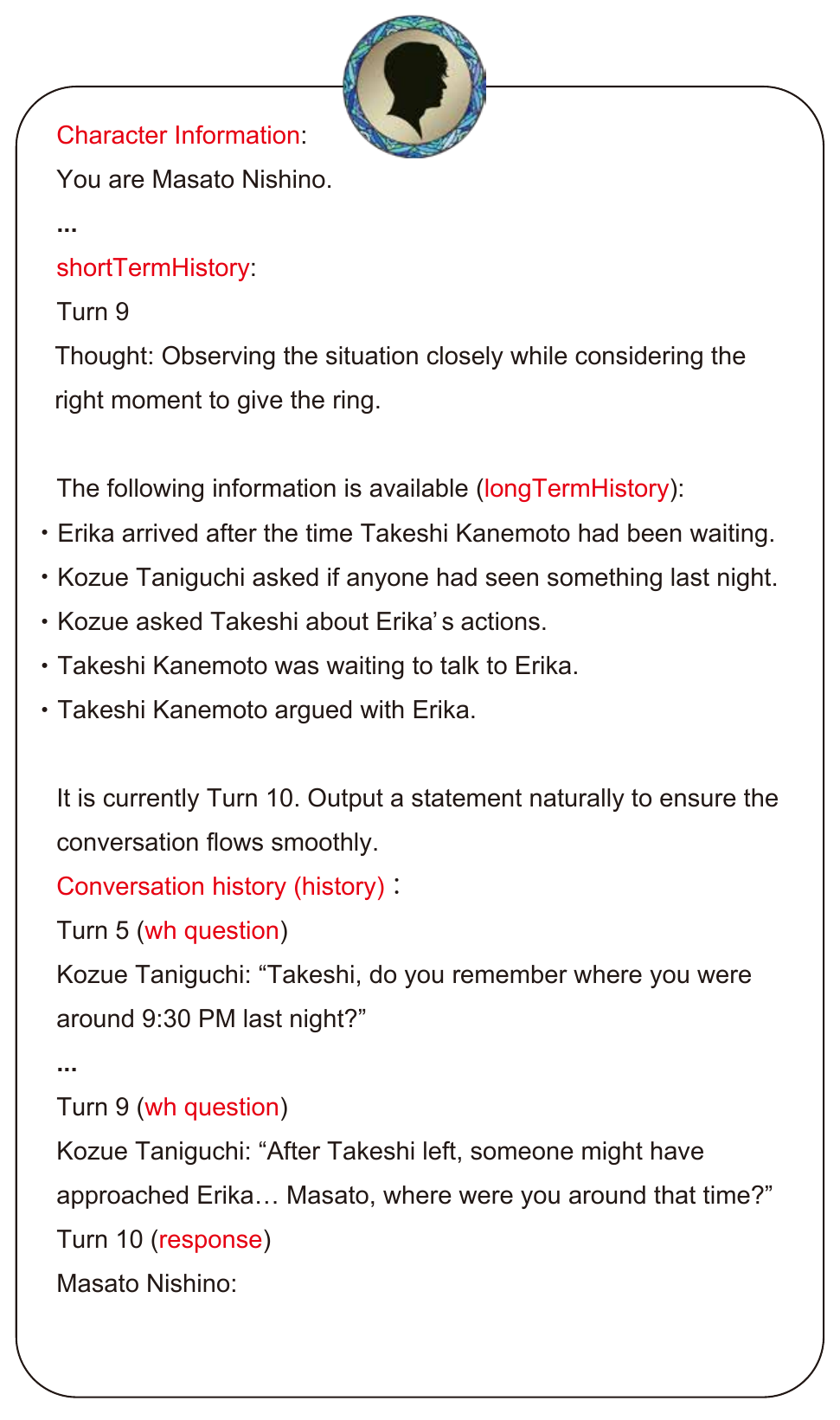}
  \caption{Example of prompt for speak().}
  \label{fig:speak()}
\end{figure}

The selected agent as speaker generates an utterance using the prompt shown in Figure \ref{fig:speak()}. This prompt consists of the character data shown in Figure \ref{fig:character information} and the three types of memory (History, shortTermHistory, longTermHistory) explained in Section \ref{sec:memory}. This enables natural utterances that consider the agent's personality, past conversation content, and policies.
\subsubsection{detectDesignation()}
\label{sec:detectDesignation()}
\begin{figure}[tbp]
  \centering
  \includegraphics[scale=0.6]{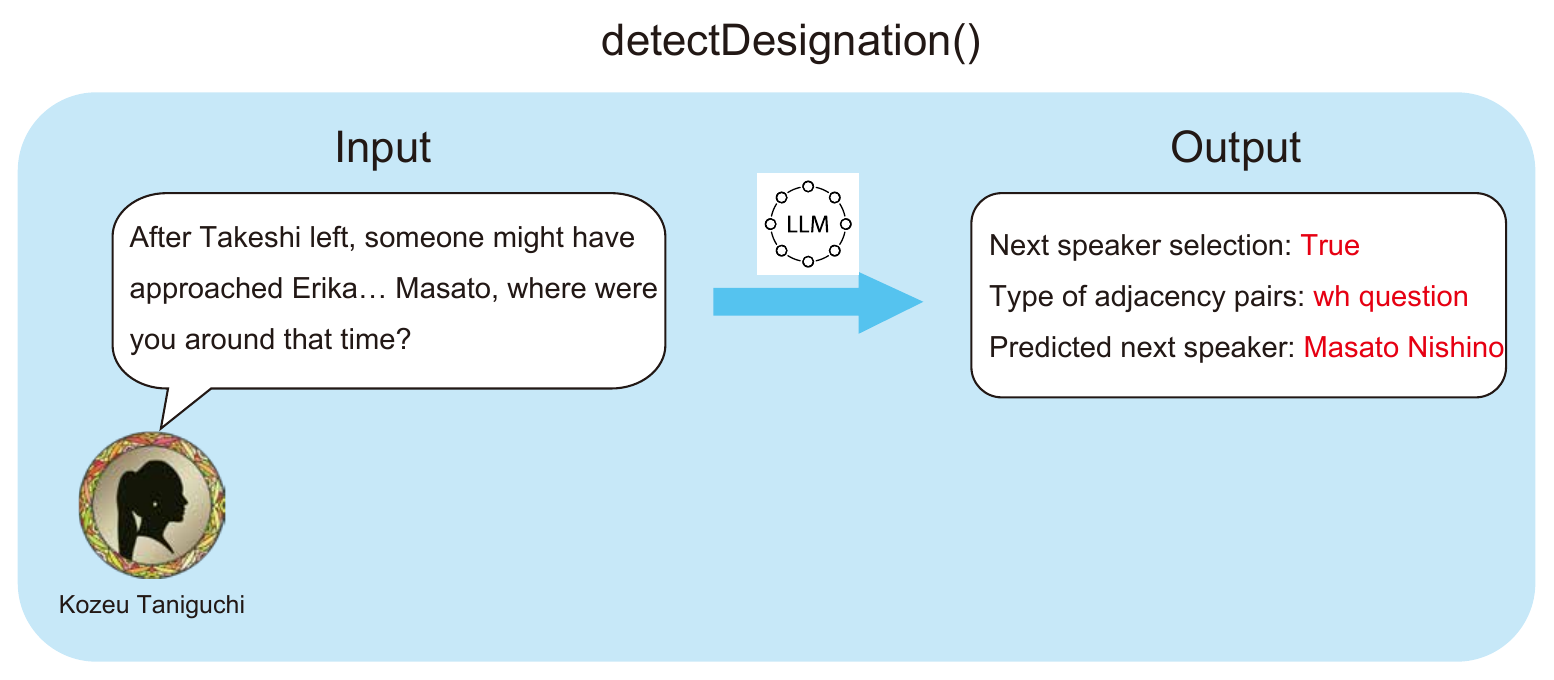}
  \caption{Example of detectDesignation().}
  \label{fig:detectDesignation()}
\end{figure}

detectDesignation() is a mechanism that detects whether the current speaker has explicitly designated the next speaker. This process uses the LLM to determine if a first pair part of an adjacency pair is present in the previous turn's utterance. When a first pair part is detected, it simultaneously classifies its type (Yes/No question, addressing, etc.) and estimates the agent addressed by the utterance. In the example shown in Figure \ref{fig:detectDesignation()}, Kozue Taniguchi asks Masato Nishino ``Where were you at that time?''. When this utterance is input to detectDesignation(), the LLM outputs the detected type of first pair part (wh question) and the predicted next speaker (Masato Nishino).

Then, by incorporating the type of the corresponding second pair part into the prompt used in the following speak(), the agent designated as the next speaker is obligated to respond to the previous turn's utterance. For example, in Turn 10 of the conversation history in Figure \ref{fig:speak()}, a constraint  of ``(response)'' is imposed on the next speaker's utterance, because the previous utterance was a first pair part (wh question). This achieves coherency in adjacent utterances while maintaining natural conversation flow.

\section{Experiments and Evaluations}
\label{sec:Experiments and Evaluations}
\subsection{Experiments}
\label{sec:Experiments}
To validate the effectiveness of the proposed MMAgents, we conducted conversational simulations using a commercially available murder mystery scenario titled ``The Ghost Island Murder Case'' \cite{Boureitou}. This scenario was selected because it features characters with well-defined roles and positions, while maintaining a moderate difficulty level for non-murderer characters, with logical deductions that are challenging yet solvable.
``The Ghost Island Murder Case'' begins with a story of former college tennis team members reuniting on an isolated island after three years. The scenario features the following four characters:
\begin{itemize}
\item Kozue Taniguchi (female): A boyish character with a straightforward personality.
\item Masato Nishino (male): An energetic character. Endearing, but sometimes fails to read the room.
\item Yukiko Shiraishi (female): A caring, big-sister type character in the group, though she has a tendency to overthink.
\item Takeshi Kanemoto (male): A sincere character despite his flashy appearance.
\end{itemize}

While the scenario consists of multiple phases (exploration phase for information gathering, private conversation phase, discussion phase, reasoning phase, etc.), our experiment focused solely on the discussion phase. This choice was primarily motivated by our aim to evaluate the effectiveness of MMAgents' core functionality: human-like turn-taking. We determined that the discussion phase, with its active dialogue and exchange of opinions between participants, would be optimal for assessing the performance of our proposed method.

In our experiments, we employed multiple large language models. GPT-4o was utilized for detectDesignation() and speak(), as these tasks require sophisticated context understanding and natural speech generation. Conversely, GPT-3.5-turbo was employed for simpler tasks such as knowledge normalization (longTermHistory) and think() to optimize computational costs. To accommodate the input token limitations of LLMs, we set the retained turns for History and shortTermHistory to five turns, while longTermHistory was configured to select the top five entries based on similarity scores.

To evaluate the proposed method, we conducted experiments under the following three conditions:
\begin{description}
  \item EQUAL: The participants have equal opportunity to speak.
  \item SS: The next speaker always Selects Self.
  \item CSSN-or-SS: Current Speaker Selects Next, otherwise the next speaker Selects Self.
\end{description}
In the EQUAL condition, the order of speaking is randomly determined each round. This ensures that the number of each participant's utterances is equal, while avoiding potential order effects. In the CSSN-or-SS condition, the turn-taking system described in Section \ref{sec:Turn-taking System} determines the speaking order. The SS condition is the same as the CSSN-or-SS condition except that it does not have the detectDesignation() mechanism, which is used for speaker selection in the next turn.

For each condition, we generated 50 sets of 10-turn conversations. The results were then evaluated using the evaluation methods described in the following subsection, enabling a statistical analysis of the effectiveness of our proposed approach.

\subsection{Evaluations}
\label{sec:Evaluations}
To evaluate the conversations generated by MMAgents, we adopted the following three approaches:
\begin{enumerate}
\item Analysis of dialogue breakdown: To assess the naturalness of generated conversations, we employed LLMs to analyze and evaluate the number of utterances that led to dialogue breakdowns \cite{Higashinaka2022}.
\item LLM-as-a-Judge: We defined three metrics---coherence, cooperation, and conversational diversity---and evaluated them using score-based LLM as the judging methodology \cite{Li2024,Kocmi2023}.
\item Human evaluation: We established original evaluation criteria focusing on murder mystery game progression and information sharing between agents. These criteria comprehensively assess the agents' reasoning capabilities and information-gathering abilities through analysis of conversations generated by MMAgents.
\end{enumerate}

\subsubsection{Analysis of Dialogue Breakdown}
\label{sec:Analysis of Dialogue Breakdown}
\begin{figure}
  \centering
  \includegraphics[scale=0.6]{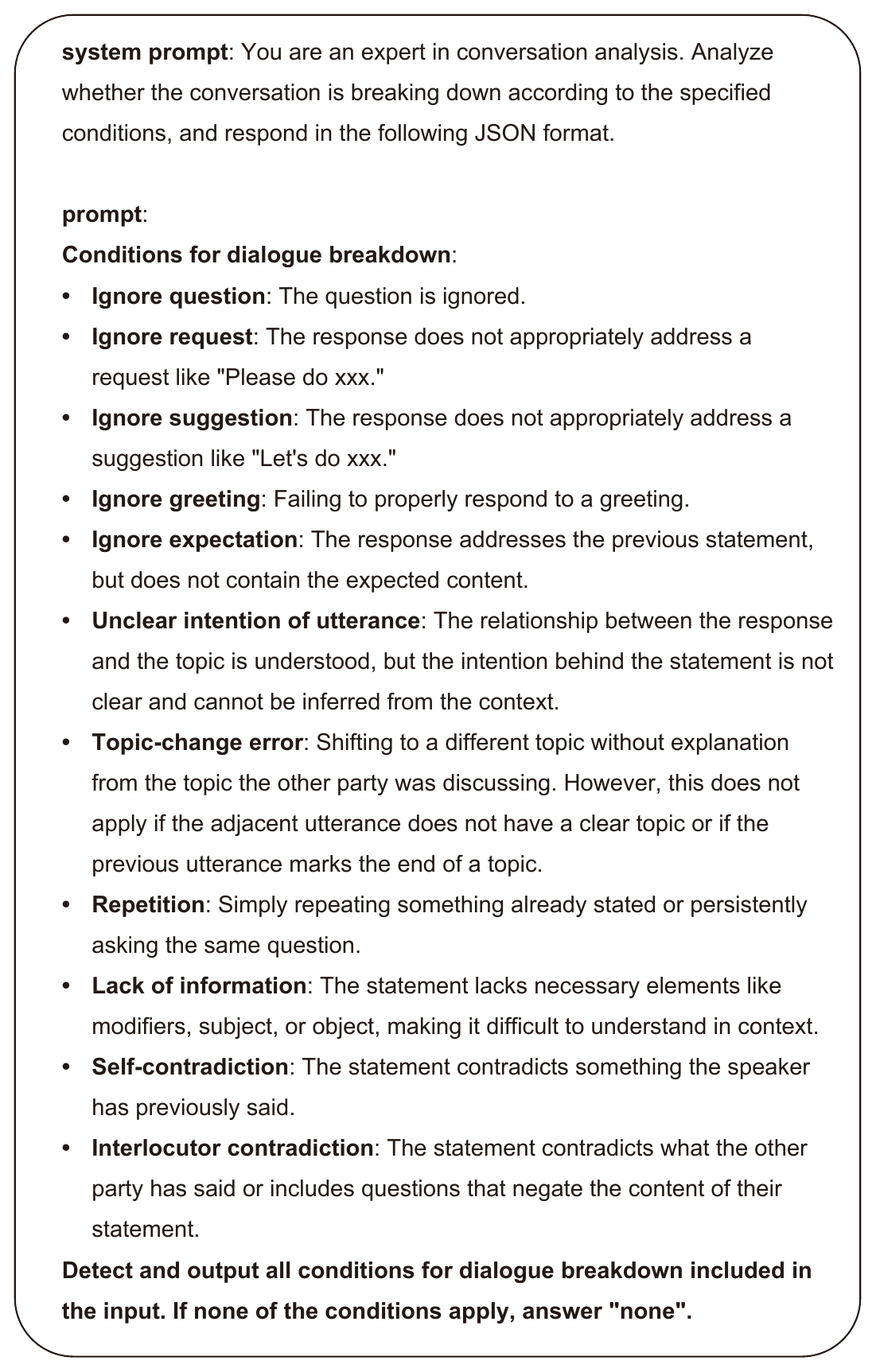}
  \caption{Prompt for the analysis of dialogue breakdown.}
  \label{fig:prompt DB}
\end{figure}

Evaluating conversational naturalness is crucial, but difficult to achieve. The evaluation of naturalness is inherently subjective, heavily dependent on evaluators' perspectives and prior experiences. Even when different evaluators assess the same conversation, their evaluations may not align, making it difficult to establish standardized evaluation criteria. Therefore, rather than directly evaluating conversational naturalness, our research adopts an indirect approach by evaluating the degree of dialogue breakdown. Specifically, we employ the ``classification of utterances that lead to dialogue breakdowns'' proposed in dialogue systems research \cite{Higashinaka2022}. Among these types, we use LLMs to analyze items corresponding to response and context-level errors shown in Table~\ref{tab:classification}.
For the analysis, we input 10-turn conversation samples generated by MMAgents into the LLM, which then identifies utterances corresponding to the categories in Table \ref{tab:classification} as breakdown utterances (B) and others as non-breakdown utterances (NB). This process is repeated 50 times, and then conversational naturalness is quantitatively evaluated through statistical analysis of the distribution of utterances identified as B. We utilized GPT-4 for this analysis, with the prompt shown in Figure \ref{fig:prompt DB}.
\begin{table}
\centering
\caption{Classification of Utterances that Lead to Dialogue Breakdowns. \cite{Higashinaka2022}}
\begin{tabular}{|l|l|l|}
\hline
\multicolumn{3}{|c|}{\textbf{Classification}} \\
\hline
& \textbf{Form} & \textbf{Content} \\
\hline
\textbf{Response} & Ignore question & Ignore expectation \\
& Ignore request & \\
& Ignore suggestion & \\
& Ignore greeting & \\
\hline
\textbf{Context} & Unclear intention of utterance & Self-contradiction \\
& Topic-change error & Interlocutor Contradiction \\
& Lack of information & Repetition \\
\hline
\end{tabular}
\label{tab:classification}
\end{table}
\subsubsection{LLM-as-a-Judge}
\label{sec:LLM as a judge}
A new approach called ``LLM-as-a-Judge'' has emerged for evaluating natural language processing tasks \cite{Li2024,Kocmi2023,Gao2023,Chiang2023}. This rapidly evolving methodology is increasingly being recognized as an alternative to traditional human evaluator-dependent methods.
The fundamental concept of the ``LLM-as-a-Judge'' approach involves inputting some text or conversation to be evaluated into LLMs and having them perform evaluations based on specific criteria or metrics. The primary advantage of this method lies in its ability to analyze large volumes of data efficiently and consistently without requiring human evaluators.

We employ LLMs to evaluate the quality of generated conversations using three metrics: coherence, cooperativeness, and diversity. Coherence evaluates the logical flow and absence of contradictions in conversations, with scores ranging from 1 (contradictory and illogical) to 5 (consistent and logical). Cooperativeness evaluates how collaboratively participants engage in information exchange and solving problems, with scores ranging from 1 (uncooperative) to 5 (cooperative). Conversational diversity evaluates the absence of repetitive content and the presence of varied opinions and perspectives, with scores ranging from 1 (no diversity) to 5 (high diversity). Coherence indicates the logical flow of conversation, Cooperativeness reflects the quality of participant interactions, and diversity represents the richness and depth of the conversation.
In the evaluation process, each conversation sample is input into the LLM, which outputs scores from 1 to 5 for each of the three metrics mentioned above. We utilized GPT-4 for this evaluation.

\subsubsection{Human evaluation}
\label{sec:Human judge}
\begin{figure}
  \centering
  \includegraphics[scale=0.5]{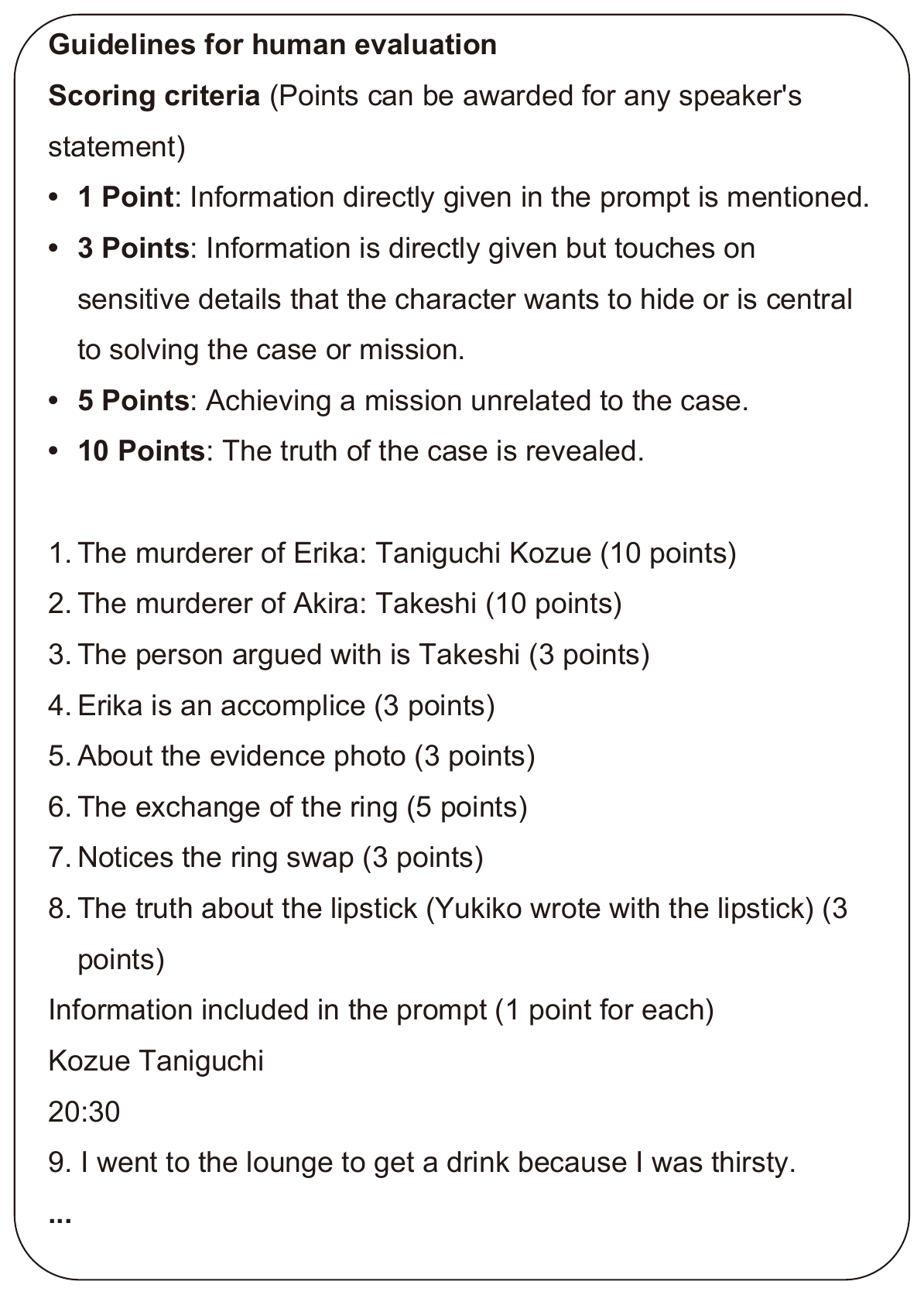}
  \caption{Guidelines for human evaluation.}
  \label{fig:HJ Guideline}
\end{figure}

To evaluate the quality and effectiveness of conversations in the Murder Mystery scenarios, the authors developed original evaluation criteria and conducted detailed evaluations of each conversation from the perspectives of information-sharing efficiency and discussion progression. Our evaluation criteria were designed based on the hypothesis that smooth conversation facilitates logical discussion, ultimately leading to the game's objective of solving the case. A portion of these evaluation criteria is shown in Figure \ref{fig:HJ Guideline}.

The evaluation of information-sharing efficiency measures the activity of information exchange, which forms the foundation for in-depth discussion. Specifically, points are awarded when character-specific information is appropriately disclosed during conversation. This quantitatively evaluates the quality of information sharing that serves as the basis for case-solving reasoning.

The evaluation of discussion progression measures the development of reasoning based on shared information and the progress toward solving the case. Points are awarded when characters demonstrate logical reasoning and insights, or when significant facts are revealed. This enables quantitative evaluation of progress toward the task of uncovering the truth behind the case.

This methodology enables systematic evaluation of the entire process, from information sharing through logical reasoning to case resolution. In particular, by considering the specific characteristics of murder mysteries, we can more concretely verify the effectiveness of our proposed method.
\section{Results}
\label{sec:Results}
\begin{figure}[tbp]
  \centering
  \includegraphics[width=\linewidth]{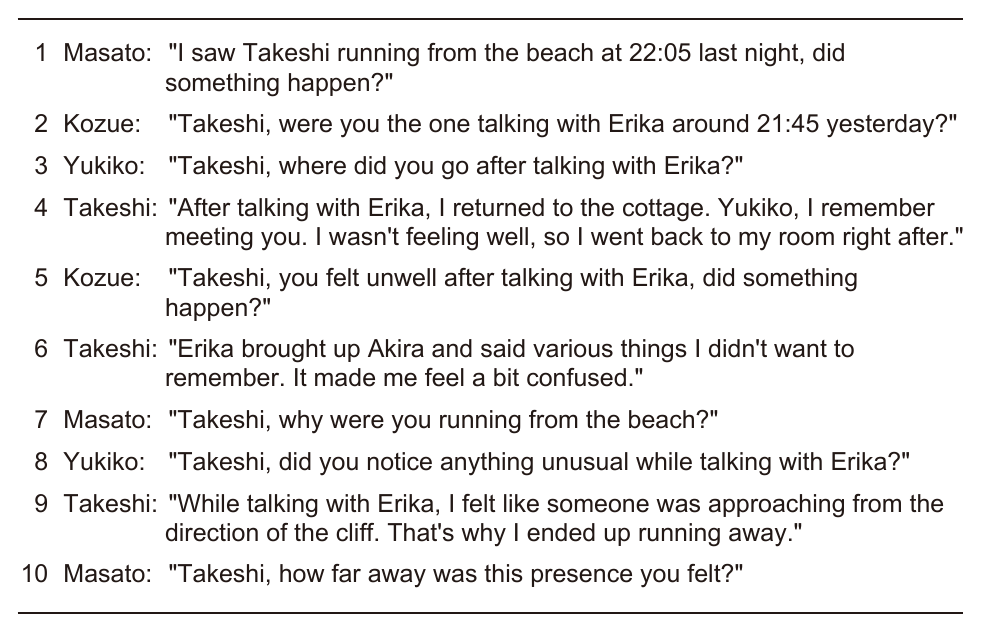}
  \caption{Example conversation in EQUAL condition.}
  \label{example:EQUAL}
\end{figure}

\begin{figure}[tbp]
  \centering 
  \includegraphics[width=\linewidth]{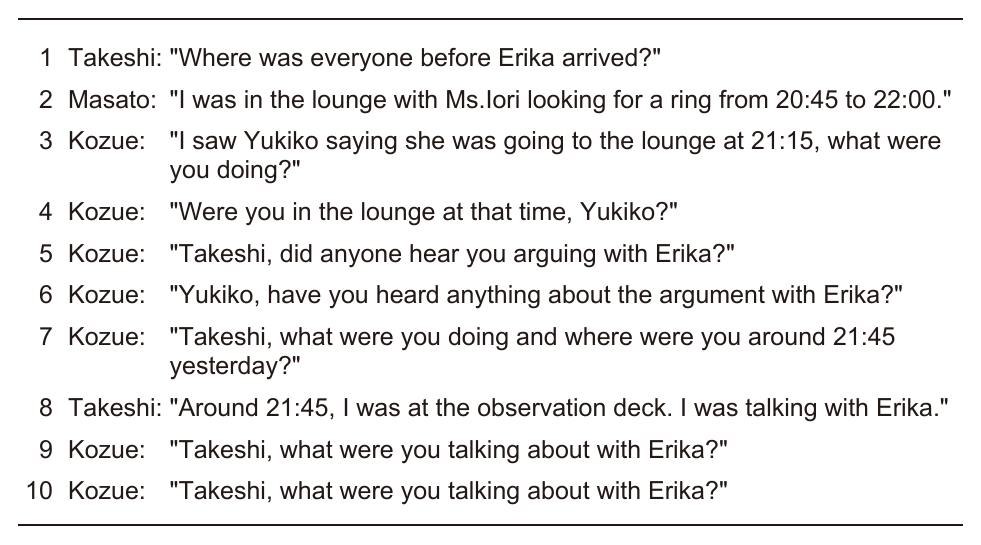}
  \caption{Example conversation in SS condition.}
  \label{example:SS}
\end{figure}

\begin{figure}[tbp]
  \centering
  \includegraphics[width=\linewidth]{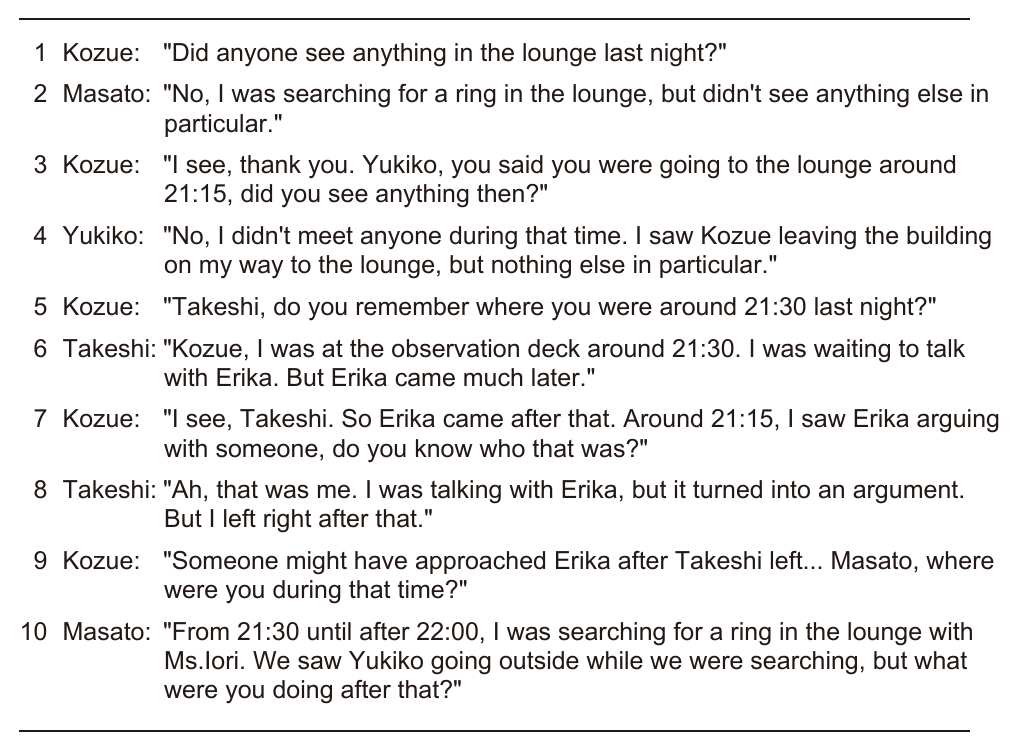}
  \caption{Example conversation in CSSN-or-SS condition.}
  \label{example:CSSN-or-SS}
\end{figure}

We compared three types of generated conversations: those with equal speaking turns and opportunities (EQUAL) as detailed in Section \ref{sec:Experiments}, those generated using only the Self-Selection mechanism (SS), and those generated using our proposed approach incorporating the Current Speaker Selects Next mechanism (CSSN-or-SS). Examples of generated conversations are shown in Figures \ref{example:EQUAL}, \ref{example:SS}, and \ref{example:CSSN-or-SS}.

Examining the EQUAL condition example in Figure \ref{example:EQUAL}, in Turn 1, Masato asks Takeshi ``Did something happen?'' In Turn 2, Kozue follows with another question to Takeshi. Turn 3 similarly contains a question directed at Takeshi. Finally, in Turn 4, Takeshi responds to these questions. Thus, in the EQUAL condition, when an utterance that corresponds to the first pair part of an adjacency pair, such as addressing someone or asking a specific question, occurs, the generation of an utterance that corresponds to the second pair part requires waiting until one's turn comes around, which may result in inefficient conversation.

In the SS condition example shown in Figure \ref{example:SS}, Turn 1 shows Takeshi asking all participants ``Where was everyone?'' In Turn 2, Masato explains his behaviors in response to the Turn 1 question. In Turn 3, Kozue asks Yukiko ``What were you doing?'' However, from Turn 4 to Turn 7, Kozue continues to ask questions. This pattern of consecutive questions from the same speaker without consideration for the second pair part of an adjacency pair was frequently observed. This is due to such an agent with high importance scores monopolizing turns.

In the CSSN-or-SS condition example shown in Figure \ref{example:CSSN-or-SS}, Turn 1 shows Kozue asking all participants ``Did anyone see anything in the lounge last night?'' In Turn 2, Masato responds to Turn 1's question ``I was in the lounge but didn't see anything.'' In Turn 3, Kozue uses a sequence-closing third \cite{Schegloff2007} saying ``I see, thank you'' to conclude the conversational sequence with Masato. Kozue then addresses Yukiko, asking a similar question based on memory that Yukiko had said she would ``go to the lounge.'' In Turn 4, Yukiko provides the second pair part of the adjacency pair, responding ``I didn't meet anyone at that time.'' In subsequent turns, the conversation flow continues with Kozue addressing specific participants with questions and appropriate agents providing responses.

\begin{figure}[tbp]
  \centering
  \includegraphics[scale=0.5]{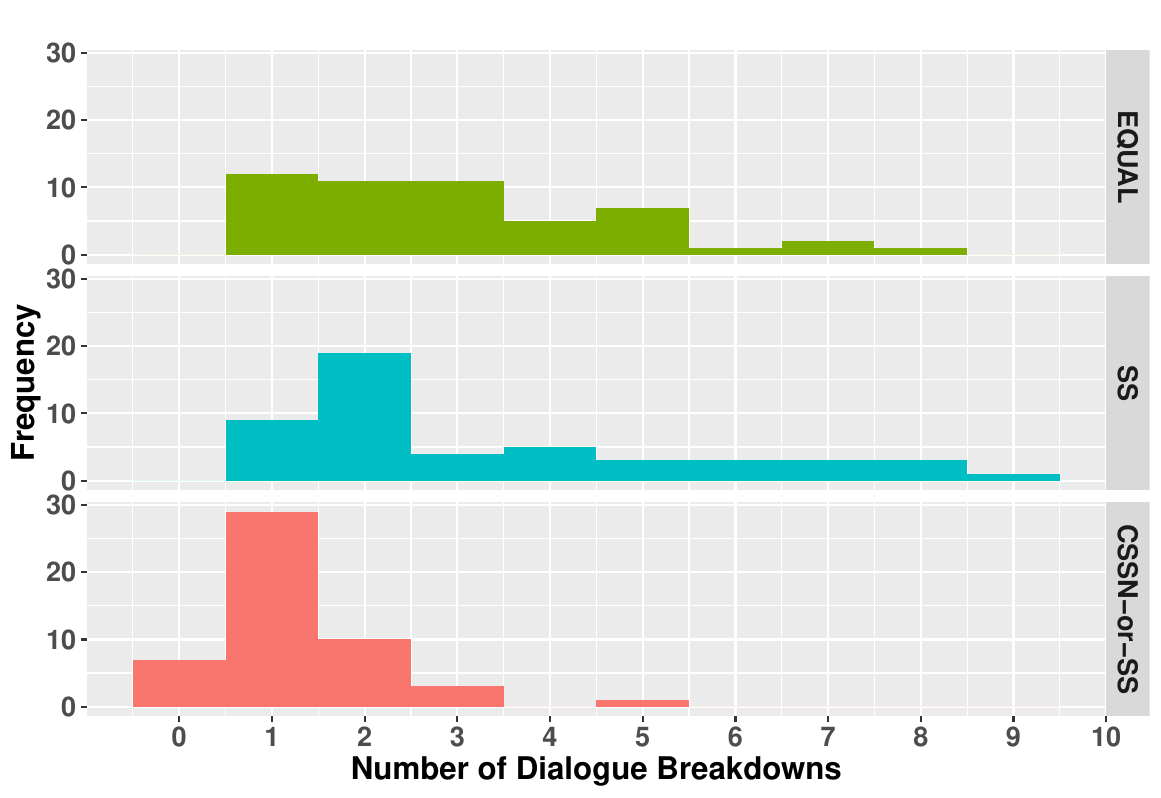}
  \caption{The number of utterances that lead to dialogue breakdowns within 10 turns.}
  \label{fig:DB}
\end{figure}

Figure \ref{fig:DB} shows the analysis results of dialogue breakdown described in Section \ref{sec:Analysis of Dialogue Breakdown}. In both the EQUAL and SS conditions, the number of utterances that led to dialogue breakdown per 10 turns showed a wide distribution from one to eight utterances. Conversely, the CSSN-or-SS condition showed a narrow distribution centered around one utterance. Kruskal-Wallis testing revealed significant differences between conditions ($\chi^2 = 42.171$, $p < 0.001$). Dunn's multiple comparison test (with Bonferroni correction) showed that the CSSN-or-SS condition significantly reduced utterances that led to dialogue breakdowns compared to the EQUAL condition ($p < 0.001$) and the SS condition ($p < 0.001$).

\begin{figure}[tbp]
  \centering
  \begin{minipage}{\textwidth}
      \centering
      \includegraphics[scale=0.5]{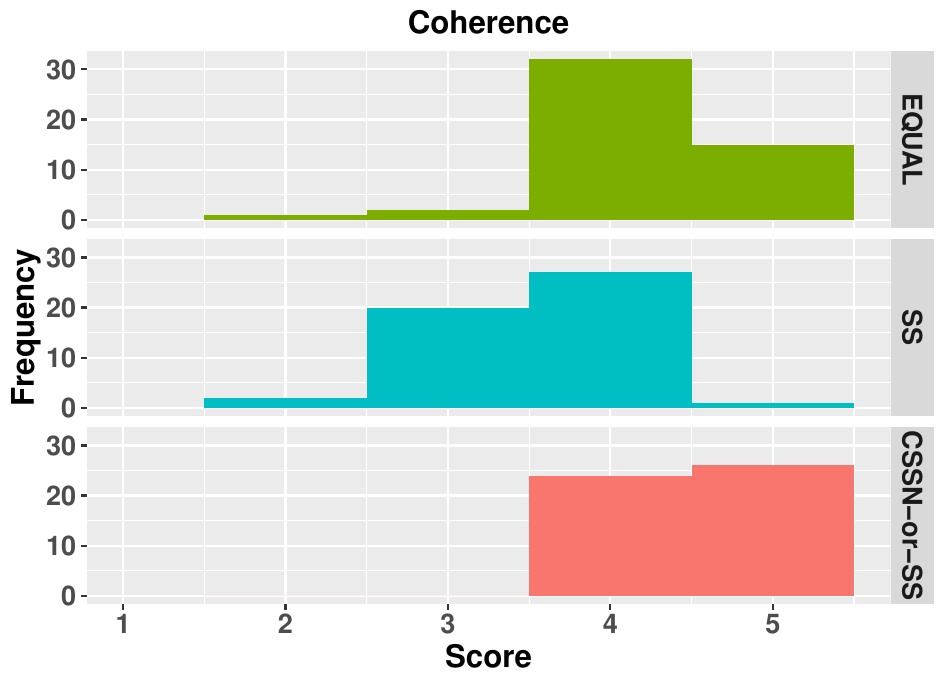}
  \end{minipage}
  \hfill
  \begin{minipage}{\textwidth}
      \centering
      \includegraphics[scale=0.5]{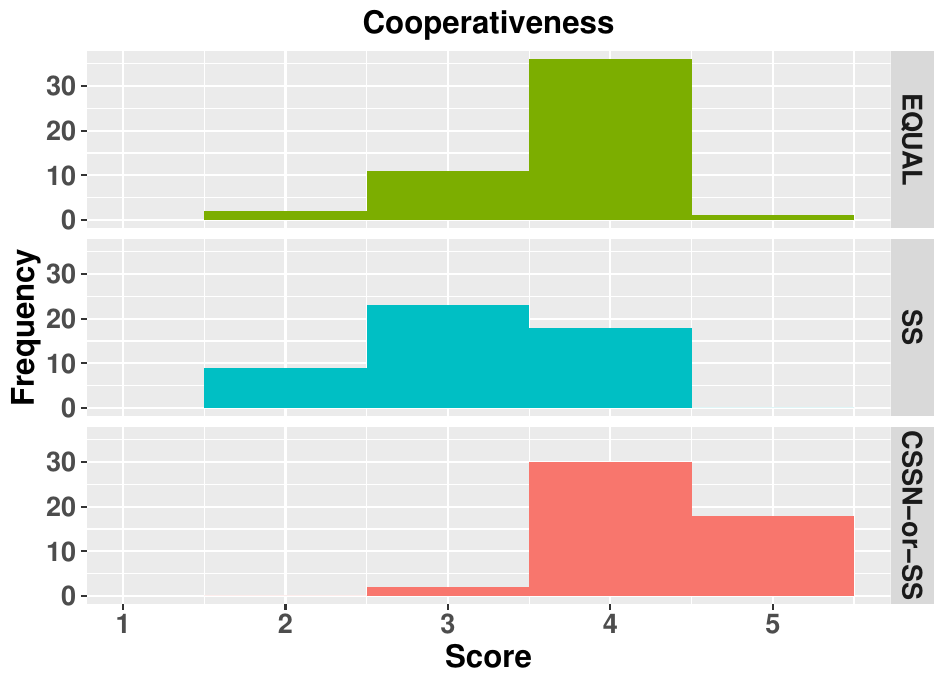}
  \end{minipage}
  \hfill
  \begin{minipage}{\textwidth}
      \centering
      \includegraphics[scale=0.5]{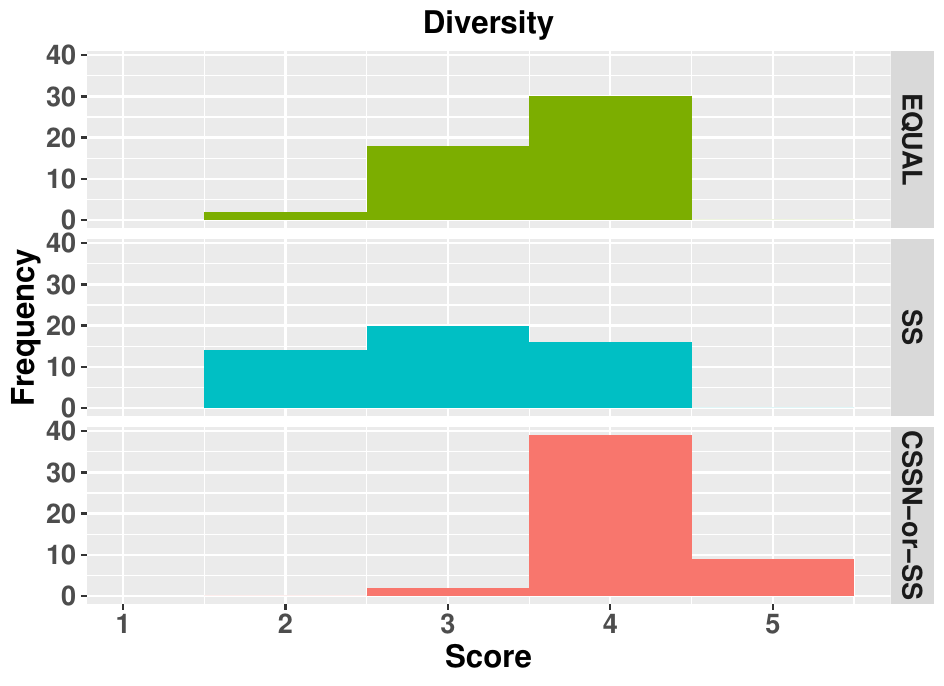}
  \end{minipage}
  
  \caption{Result of LLM judge.}
  \label{fig:LLM as a judge}
\end{figure}

Figure \ref{fig:LLM as a judge} shows the LLM-as-a-Judge evaluation results described in Section \ref{sec:LLM as a judge}. For the metrics of coherence, cooperativeness, and diversity, the EQUAL condition showed peaks at score 4, while the SS condition showed wide distributions from scores 2 to 4. The CSSN-or-SS condition distributed across scores 4 and 5, with diversity showing a notable peak at score 4. Kruskal-Wallis testing revealed significant differences between conditions for all metrics (coherence: $\chi^2 = 51.784$, $p < 0.001$; cooperativeness: $\chi^2 = 56.718$, $p < 0.001$; diversity: $\chi^2 = 52.973$, $p < 0.001$). Dunn's multiple comparison test (with Bonferroni correction) showed no significant difference between CSSN-or-SS and EQUAL conditions for coherence ($p = 0.084$), but significant differences between all other condition pairs ($p < 0.01$). For cooperativeness and diversity, significant differences were found between all condition pairs ($p < 0.01$).

\begin{figure}[tbp]
  \centering
  \includegraphics[width=.7\hsize]{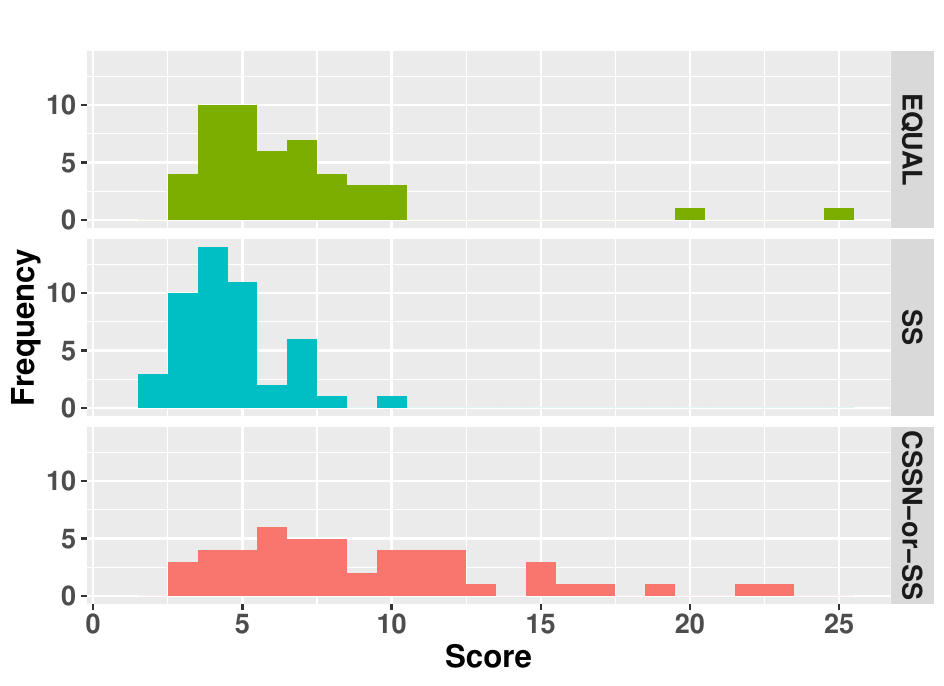}
  \caption{Result of Human evaluation.}
  \label{fig:Handcrafted}
\end{figure}

Finally, Figure \ref{fig:Handcrafted} shows the results of human evaluation. The CSSN-or-SS condition promoted appropriate question-answer exchanges and information sharing, showing a wide distribution centered around score 8. In contrast, the EQUAL and SS conditions exhibited dialogue breakdowns such as monopolization of speech by specific agents and consecutive questioning, resulting in narrow distributions concentrated around score 4. Kruskal-Wallis testing revealed significant differences between conditions ($\chi^2 = 40.644$, $p < 0.001$). Dunn's multiple comparison test (with Bonferroni correction) showed significant differences between all condition pairs ($p < 0.01$).

\section{Discussion}
\label{sec:Discussion}
\begin{figure}[tbp]
  \centering
  \includegraphics[width=\hsize]{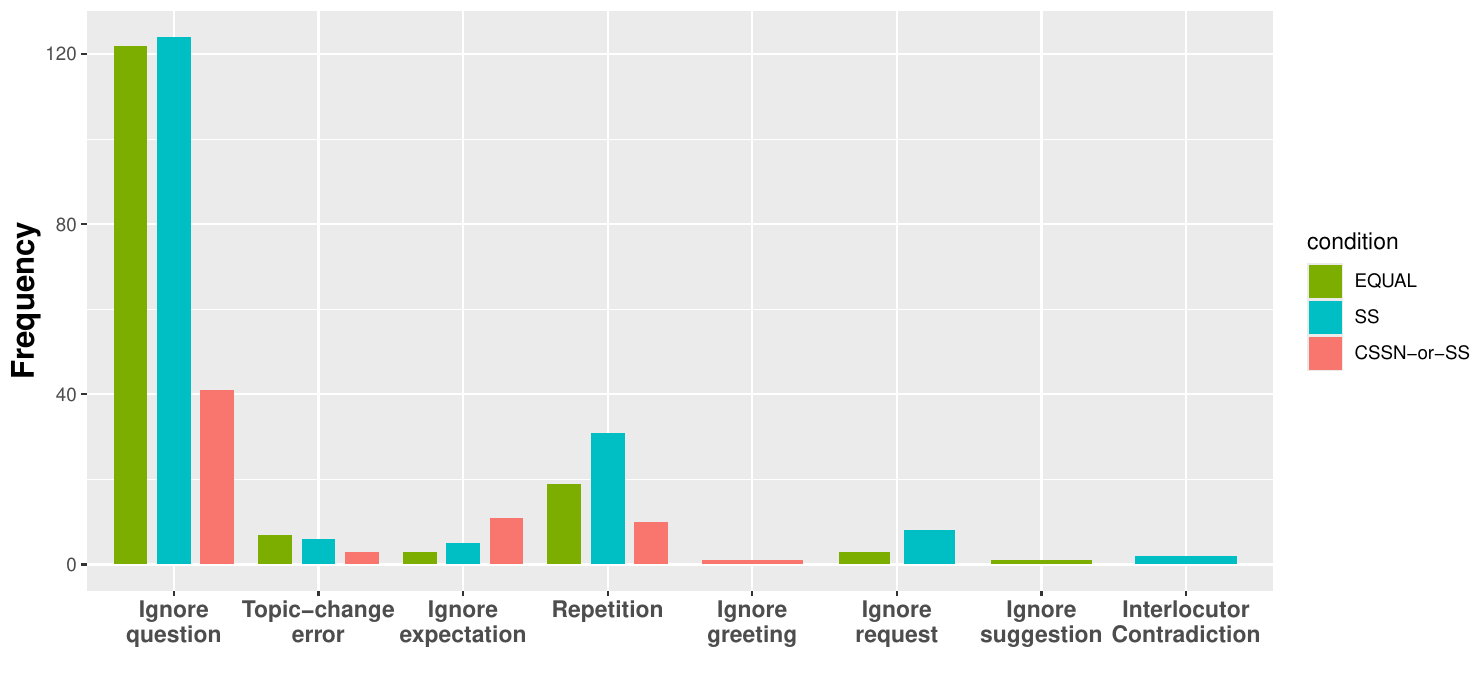}
  \caption{The classified types of utterances that lead to dialogue breakdowns}
  \label{fig:Classification}
\end{figure}

\begin{figure}[tbp]
  \centering
  \includegraphics[width=\linewidth]{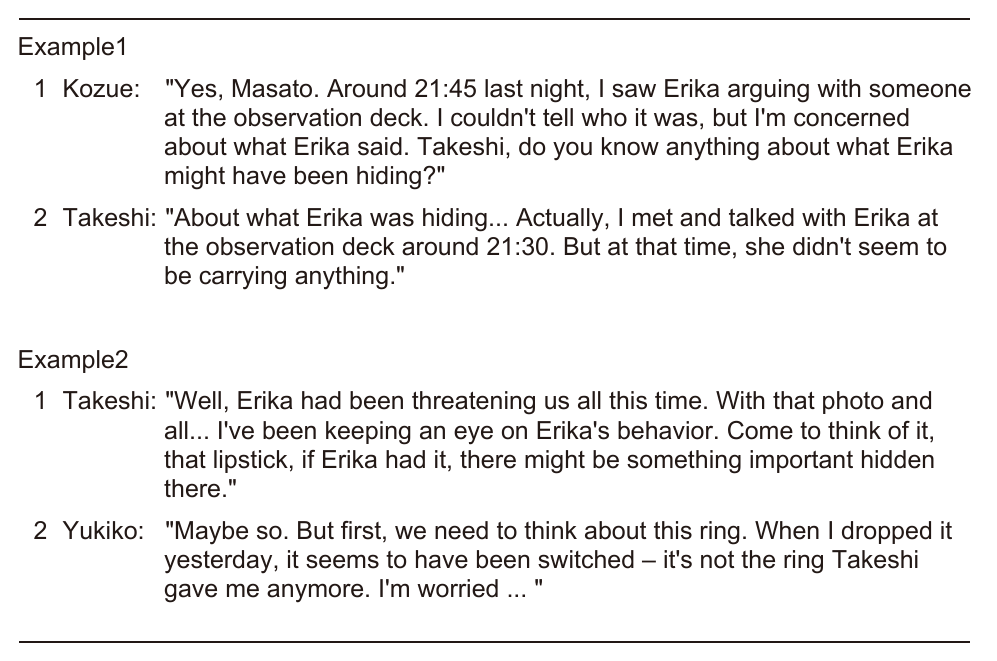}
  \caption{Example of Ignore expectation.}
  \label{example:ignore_expectation}
\end{figure}

The experimental results of this study clearly demonstrate that the next-speaker selection mechanism utilizing adjacency pairs in turn-taking systems improves the quality of multi-party conversations in multiple aspects. 
From the analysis of dialogue breakdowns, a significant decrease in the number of utterances that led to dialogue breakdowns was observed.
Figure \ref{fig:Classification} shows the frequency distribution of classified dialogue breakdown types (refer to Table \ref{tab:classification}) under each condition. In the CSSN-or-SS condition, a notable decrease in ignoring the question was confirmed compared to both the EQUAL and SS conditions. This is considered to be due to the next-speaker selection mechanism clarifying response obligations for specific participants, thereby suppressing inappropriate next speaker and responses to questions. Additionally, it is suggested that by structuring the flow of dialogue and promoting responses related to previous utterances, the Current Speaker Selects Next mechanism reduced abrupt topic changes (Topic-change error) and repetition.
While approximately 40 instances of ignoring the question were identified in the CSSN-or-SS condition, detailed analysis of their content revealed characteristic patterns in addition to typical ignoring the question (e.g., cases where an agent with a response obligation asks a new question without answering).
First, a tendency was observed where responses addressed only part of the question while avoiding core information. For example, as shown in Example 1 in Figure \ref{example:ignore_expectation} where Kozue asked ``Do you know anything about what Erika might have been hiding?'', Takeshi explained the circumstances of interaction with Erika but avoided addressing the essential answer about what was being hidden.

Second, some patterns were observed where agents intentionally shifted to different topics to avoid expected responses. As shown in Example 2 in Figure \ref{example:ignore_expectation}, despite expectations for discussion about Erika's lipstick, Yukiko suddenly switched to discussing rings, representing a case of avoiding responding to the original question.

These characteristics suggest that within the context of reasoning games like Murder Mystery, the Current Speaker Selects Next mechanism influences agents' information disclosure strategies. It is considered that as response obligations became clearer, agents began to control information disclosure in a more sophisticated way while avoiding simple ignoring the question to maintain their position in the game. For instance, in Example 2, as Erika's lipstick was given as information that Yukiko needed to keep secret in her character settings, the switch to the topic of rings can be interpreted as a strategic choice to protect this secret. However, it is necessary to consider the possibility that these observed behavioral patterns might be influenced by the limitations in contextual processing capabilities of the LLMs used.

The evaluation results from the LLM-as-a-Judge demonstrate that the proposed method incorporating the Current Speaker Selects Next mechanism with adjacency pairs (CSSN-or-SS condition) comprehensively improved conversation coherence, cooperation, and diversity compared to both the EQUAL and SS conditions. These improvements can be attributed to the following advantages of introducing adjacency pairs:
First, the generation of appropriate responses to questions was promoted, enabling logical conversation development. Second, clear turn-taking encouraged active participation in information exchange and problem-solving. Third, the repetition of identical utterances was suppressed, enabling the presentation of opinions from diverse perspectives.

However, it is noteworthy that no significant difference was observed between the CSSN-or-SS condition and EQUAL condition in terms of coherence evaluation. This result may be attributed to the characteristic properties of the EQUAL condition. Specifically, in the EQUAL condition, speaking opportunities are equally distributed among all conversation participants regardless of conversation content. Consequently, even when immediate response to the first part of an adjacency pair (e.g., question) is not possible in the subsequent turn, participants are guaranteed to have a speaking opportunity in later turns, enabling them to provide the second part (e.g., answer). This structural characteristic may have ensured the eventual establishment of logical conversations, albeit not immediately.

The results of human evaluation revealed that the proposed method incorporating the Current Speaker Selects Next mechanism (CSSN-or-SS condition) showed a wide distribution centered around 8 points, achieving significantly higher scores compared to other conditions. The Murder Mystery is a specialized dialogue task that requires not only natural conversation but also information sharing among multiple characters and logical reasoning based on combining this information. The results suggest that the proposed method effectively achieved these elements.

\begin{figure}
  \centering
  \includegraphics[width=\linewidth]{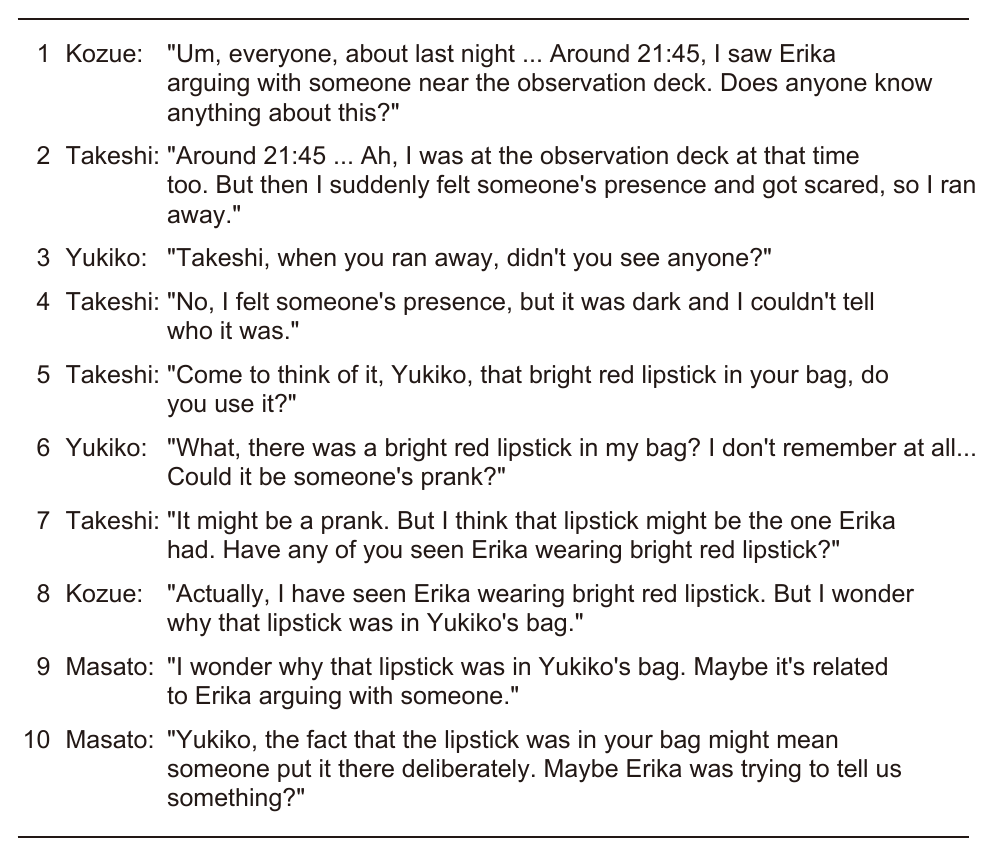}
  \caption{Example of low score conversation.}
  \label{example:low score}
  \end{figure}

However, the presence of a certain number of low scores between 4 and 6 points even in the CSSN-or-SS condition indicates that there is still room for improvement in the proposed method. Analysis of low-scoring conversation examples, as shown in Figure \ref{example:low score}, revealed a characteristic where excessive time was spent on specific topics. Specifically, in this conversation example, 6 out of 10 turns were spent on speculations about the lipstick found in Yukiko's bag, yet they did not reach the truth about the lipstick (worth 3 points). The conversation ended without sharing or reasoning about other important information, resulting in a low score. While this conversation maintains a natural flow as general dialogue, it should be improved from the perspective of ``sharing information and developing reasoning based on information,'' which is crucial in Murder Mystery games.

These analysis results suggest the importance of goal-oriented topic control. Specifically, the introduction of a mechanism that adjusts topic duration based on the importance of provided information could enable more effective reasoning processes.

Additionally, insufficient conversation turns might be one factor contributing to low scores. With fewer turns, discussions risk becoming biased toward specific topics, ending before other important information can be shared. In fact, by dedicating considerable time to speculations about the lipstick, other facts were neither shared nor verified, leaving the reasoning incomplete. To improve such situations, increasing conversation turns could potentially broaden the scope of discussion and promote the sharing and verification of crucial information related to the core of the case.

There are several limitations to the human evaluation. Particular attention should be paid to the possibility that the subjectivity of the authors themselves, as evaluators, may have influenced the results. Therefore, future work should incorporate measures to improve evaluation objectivity and reliability, such as including evaluations from multiple evaluators.

\section{Conclusion}
In this study, we implemented and verified the effectiveness of turn-taking systems, such as adjacency pairs discovered in conversation analysis research, in multi-party conversations among LLM-based agents. Based on Schegloff's theory that ``in conversational turn-taking systems, the organization of utterance sequences, such as adjacency pairs, is the source of conversational coherence'' \cite{Schegloff1990}, we aimed to achieve more natural and coherent conversations by applying these norms to interactions between AI agents.

The experimental results strongly supported this theoretical prediction. The introduction of a turn-taking system using response obligations to the first pair part of an adjacency pair significantly reduced dialogue breakdowns, improved conversational cooperation and diversity, and enhanced agents' information sharing capabilities and reasoning abilities. In particular, the next-speaker selection mechanism based on adjacency pairs enabled smooth transitions of utterances between agents and promoted the generation of contextually appropriate responses. These results demonstrate that the norms of speech communication observed in human conversations also play a crucial role in conversations between AI agents.

However, several challenges remain in this research. The current system faces difficulties in maintaining memory using longTermMemory in extended dialogues of around 30 turns, and exhibits issues with topic management between agents, leading to topic deviation. Specific examples and detailed conversation logs are available on the project's \href{https://github.com/NONO-111/LLM-Based-Multi-Conversational-Agent}{website}. Furthermore, future challenges include implementing a concept of time in conversation, such as the gradual prediction of transition-relevance places (TRPs) \cite{Sacks1974} and controlling barge-in at non-TRPs, particularly in cases where listeners seek clarification, request additional explanation, raise questions, or express counterarguments during ongoing utterances.

Moving forward, we will address these challenges and further explore the applicability of conversation analysis theory in dialogues between AI agents. In particular, based on insights gained from the analysis of conversation data, we plan to improve the long-term memory mechanism and refine topic management.

\bibliographystyle{elsarticle-num} 
\bibliography{reference.bib}
\end{document}